%% file: cvpr22.tex
\documentclass[10pt,twocolumn,letterpaper]{article}

\usepackage[pagenumbers]{cvpr} 

\usepackage{graphicx}
\usepackage{amsmath}
\usepackage{amssymb}
\usepackage{booktabs}
\usepackage{arydshln}

\newcommand{\beginsupplement}{%
        \setcounter{table}{0}
        \renewcommand{\thetable}{A\arabic{table}}%
        \setcounter{figure}{0}
        \renewcommand{\thefigure}{A\arabic{figure}}%
     }

\usepackage[pagebackref,breaklinks,colorlinks]{hyperref}

\usepackage[capitalize]{cleveref}
\crefname{section}{Sec.}{Secs.}
\Crefname{section}{Section}{Sections}
\Crefname{table}{Table}{Tables}
\crefname{table}{Tab.}{Tabs.}

\def\cvprPaperID{5248} 
\def\confName{CVPR}
\def\confYear{2022}

\title{A Unified Architecture of Semantic Segmentation and Hierarchical Generative Adversarial Networks for Expression Manipulation\\[.75ex] 
  {\normalfont\large 
    Rumeysa Bodur{\textsuperscript{1}}, \normalfont\large Binod Bhattarai\textsuperscript{2}, Tae-Kyun Kim\textsuperscript{1,3}%
  }\\[-1.5ex]
}

\author{
    {%
        \textsuperscript{1}Imperial College London, UK
    }
    \and
    {%
        \textsuperscript{2}University College London, UK%
    }
    \and
    {%
        \textsuperscript{3}KAIST, South Korea%
    }
}

\begin{document}
\maketitle

\input{abstract}
\input{introduction}
\input{related_work}
\input{proposed_method}
\input{experiments}
\input{conclusion}

\input{limitations}

\clearpage

{\small
\bibliographystyle{ieee_fullname}
\bibliography{egbib}
}

\input{supplementary}

\end{document}

%% file: abstract.tex
\begin{abstract}
Editing facial expressions by only changing what we want is a long-standing research problem in Generative Adversarial Networks (GANs) for image manipulation. Most of the existing methods that rely only on a global generator usually suffer from changing unwanted attributes along with the target attributes. Recently, hierarchical networks that consist of both a global network dealing with the whole image and multiple local networks focusing on local parts are showing success. However, these methods extract local regions by bounding boxes centred around the sparse facial key points which are non-differentiable, inaccurate and unrealistic. Hence, the solution becomes sub-optimal, introduces unwanted artefacts degrading the overall quality of the synthetic images. 
Moreover, a recent study has shown strong correlation between facial attributes and local semantic regions. To exploit this relationship, we designed a unified architecture of semantic segmentation and hierarchical GANs. A unique advantage of our framework is that on forward pass the semantic segmentation network conditions the generative model, and on backward pass gradients from hierarchical GANs are propagated to the semantic segmentation network, which makes our framework an end-to-end differentiable architecture. This allows both architectures to benefit from each other. To demonstrate its advantages, we evaluate our method on two challenging facial expression translation benchmarks, AffectNet and RaFD, and a semantic segmentation benchmark, CelebAMask-HQ across two popular architectures, BiSeNet and UNet. Our extensive quantitative and qualitative evaluations on both face semantic segmentation and face expression manipulation tasks validate the effectiveness of our work over existing state-of-the-art methods.
\end{abstract}

%% file: introduction.tex
\section{Introduction}
\label{sec:intro}

\begin{figure}
    \centering
    \includegraphics[width=0.4\textwidth]{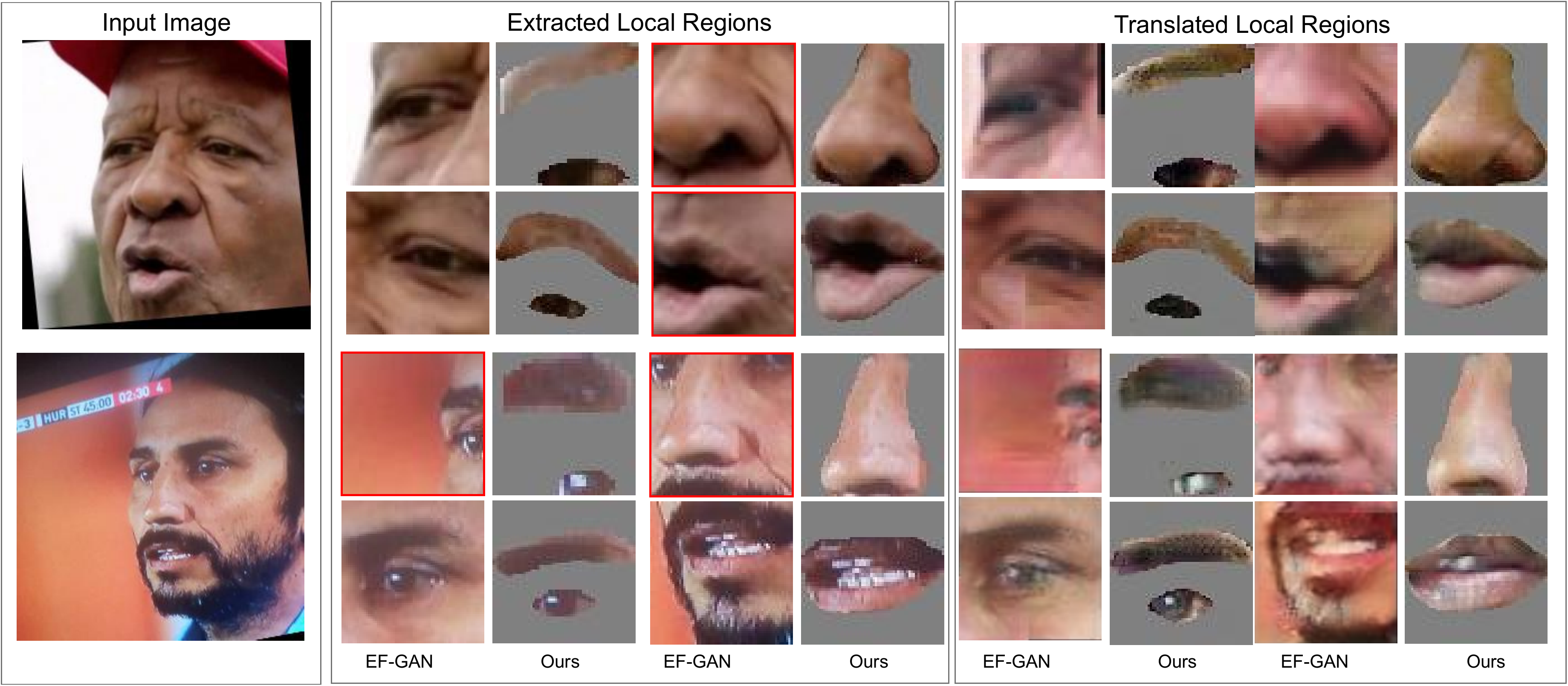}
    \caption{\textbf{A comparison of the local parts generated by the rectangular spatial division~\cite{yi2019apdrawinggan,wu2020efgan} and the proposed semantic segmentation network.} We observe that some local parts (red framed) are not extracted correctly by existing approaches. Their translation by local networks (from anger to neutral), on the right-hand side, highlight the artefacts caused by spatial division.} 
    \label{fig:motivation}
\end{figure}

Manipulating the facial expression given a single RGB image and a target expression label is a challenging and important research problem~\cite{zhu2017cyclegan,choi2018stargan,he2017attgan,bhattarai2020inducing,liu2019stgan}. 
Editing what we want precisely and accurately is another 
goal of image-to-image translation~\cite{he2017attgan}. Most  existing methods rely only on a global generator and suffer from editing unnecessary attributes in addition to related attributes~\cite{choi2018stargan,he2017attgan}.
Furthermore, modeling facial expressions only with global parameters introduces unwanted artefacts and blurriness~\cite{wu2020efgan}. This stems from the fact that global parameters cannot handle the detailed changes in local regions that occur during the switch from one expression to another. 
Recently, hierarchical architectures~\cite{yi2019apdrawinggan,wu2020efgan} have shown effectiveness in addressing such issues that are present in methods consisting only of a global network. The primary reason for the effectiveness of these hierarchical architectures is due to their extra capability to deal with specific local regions independently and to learn specialised parameters for regions of interest.


APDrawing~\cite{yi2019apdrawinggan} and EF-GAN~\cite{wu2020efgan} are two representative hierarchical architectures for facial image tasks. Their major drawback lies in the way they extract the local regions. Both APDrawing and EF-GAN extract local patches by a bounding box centred at the landmark coordinates of key local regions of faces as shown in Figure~\ref{fig:motivation}. This approach limits the shape of a semantic region to be rectangular and assumes that the regions for every sample are of equal size, whereas in reality, the shape and size of the semantic region differs with regard to the person and the expression  (Figure~\ref{fig:motivation}).
In addition, such local parts overlap with each other, which causes artefacts on the synthetic images in overlapping areas and borders of the rectangles. Moreover, such methods are not differentiable.

A recent study on semantic segmentation conditioned GANs for face attributes has identified strong tie-ups between a specific set of semantic parts and certain facial attributes~\cite{Wei_2020_ACCV}. Hence, it is essential to accurately estimate the local regions for learning attribute specific parameters. 
Inspired by the findings in~\cite{Wei_2020_ACCV} and to overcome the issues with existing hierarchical methods for facial image manipulation~\cite{yi2019apdrawinggan, wu2020efgan}, we propose a unified architecture of semantic segmentation and hierarchical GANs as shown in Figure~\ref{fig:proposed_pipeline}. 
On the forward pass, for a given RGB image, we employ the segmentation network to extract local parts by mapping each pixel to only one local region. This helps us to extract local parts that are disjoint and are of variable shape and size. 
Local parts relying on the semantic segmentation network are fed into their respective local GANs. We train all local GANs along with the global GAN to translate the given image to the target expression. 
On the backward pass, we condition the semantic segmentation network by propagating gradients from the hierarchical GANs as shown with dashed lines in Figure~\ref{fig:proposed_pipeline}.
We would like to emphasise that our method does not need any extra semantic segmentation annotations to update the parameters of the semantic segmentation network. We propose to use the tempered softmax function on the semantic segmentation network to bridge the gradients from the hierarchical GANs to the semantic segmentation network in order to make our framework end-to-end differentiable. As the accuracy on attribute manipulation is directly correlated with the precision of the local semantic parts~\cite{Wei_2020_ACCV}, with the improvement on the performance of hierarchical GANs, the precision in the semantic segmentation improves as well. 



Conditioning GANs on higher-order representations such as word2vec~\cite{reed2016texttoimage}, semantic segmentation~\cite{Chen2019CVPR}, 3D representations~\cite{nguyen2019hologan} and semantic layouts have been successfully applied before. However, to the best of our knowledge, this is the first unified framework to condition both semantic segmentation and hierarchical GANs from both directions in an end-to-end fashion.

We summarise our contributions in the following points:
\begin{itemize}
    \item We propose a unified end-to-end framework to train the semantic segmentation network and hierarchical GANs conditioning one module by another.
    \item We perform extensive evaluations on two challenging face expression manipulation benchmarks and a semantic segmentation benchmark across two architectures. Our method attains new state-of-the-art results.  
\end{itemize}

%% file: related_work.tex
\section{Related Work}
\noindent\textbf{Semantic Segmentation-Guided GANs.} 
GANs \cite{goodfellow2014gan} have achieved remarkable results in generating realistic images.
Class conditional GANs~\cite{mirza2014cgan,odena2017conditional} gained further control on generated images. 
Over the past years, semantic representation of texts~\cite{reed2016texttoimage,zhang2017stackgan}, 3DMM rendered facial images~\cite{gecer2018semi}, image layouts~\cite{zhao2019image}, graph induced face attribute representations~\cite{bhattarai2020inducing} and semantic segmentation~\cite{song2018spg} have been employed to guide the generator. 
Existing studies relying on semantic segmentation have been applied for image inpainting~\cite{song2018spg}, semantic image editing~\cite{ling2021editgan} and domain adaptation~\cite{cherian2019sem,murez2018image,tomei2019art2real}. None of these methods utilise semantic segmentation to condition local GANs.

\noindent \textbf{GAN regulated Semantic Segmentation.}
A recent study \cite{souly2017semisupss} proposes a semi-supervised framework, where fake training samples generated by GANs are passed into the discriminator along with annotated data and a large amount of unannotated data. This forces the real samples to be closer in the feature space, hence improves the semantic segmentation performance. \cite{hong2018cganstructuredss} proposes to utilise conditional GANs to close the domain gap between synthetic and real images. Similarly, a study \cite{Chen2019CVPR} exploits depth information to close this domain gap for leveraging synthetic data. ReDo \cite{chen2019segredrawing} employs GANs to redraw object segments in order to guide the segmentation process in an unsupervised manner with the idea that such perturbation does not affect realism.

\noindent\textbf{Facial Expression Manipulation.} 
Recently, conditional GANs have been widely applied to manipulate facial expressions. 
StarGAN~\cite{choi2018stargan} proposes to train both the generator and the discriminator conditioned on target labelled one-hot encoded vectors to translate the expressions. 
Similarly, GANimation~\cite{pumarola2018ganimation} proposes to utilise action units~\cite{friesen1978facial} in an encoder-decoder network.
GANs conditioned on sparse 2D face landmarks~\cite{di2018gpgan,song2018g2gan,qiao2018gcgan} have also been useful to manipulate the expressions more accurately. 
Some of the recent studies go beyond landmarks to exploit geometric information and make use of depth or surface normals~\cite{shu2017neuralface,nagano2018pagan,bodur2021densegeo}. 
This conditioning captures the overall shape information but fails to localise the specific semantic regions exclusively. To overcome these issues, we propose to rely on hierarchical GANs guided by semantic segmentation. 

\noindent\textbf{Hierarchical GANs.} 
APDrawing\cite{yi2019apdrawinggan} proposes to extract local regions based on landmark positions and trains local networks for these regions. Cascade EF-GAN\cite{wu2020efgan} employs the same approach on expression manipulation with an additional attention mechanism. Nevertheless, such an approach does not precisely catch the semantic local regions, which leads to artefacts on the regional borders.
LGGAN \cite{tang2019lggan} proposes to utilise a global image-level and local class-specific GANs guided by semantic maps for semantic image synthesis and cross-view image translation. The semantic segmentation maps are given as input and the ground truth target RGB image is available during training. We propose to employ a semantic segmentation network along with hierarchical GANs and train our framework in an end-to-end manner where both tasks benefit from each other. 

%% file: proposed_method.tex
\section{Proposed Method}

\begin{figure*}
    \centering
    \includegraphics[trim= 0cm 0cm 0cm 0cm, clip, width=0.7\linewidth]{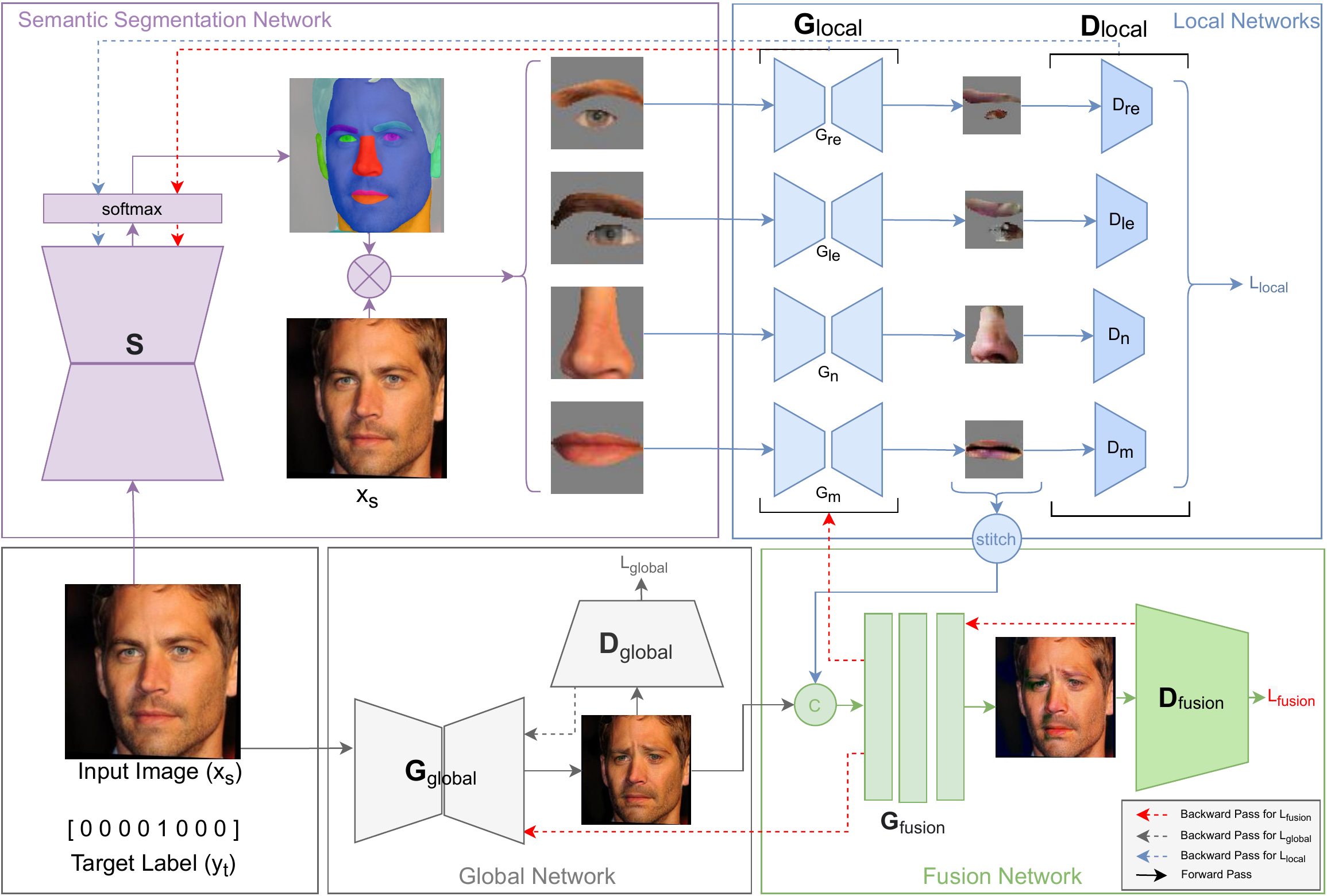}
    \caption{\textbf{Pipeline of the proposed method.} Our pipeline consists of a global network, semantic segmentation network, fusion network and four local networks. The global network takes the whole image as input and aims to capture the global structure while the local networks operate on the respective semantic regions. These regions are extracted by utilising a semantic segmentation network. The generated local images are stitched together and passed to the fusion network after being concatenated with the global output. $\otimes$ and \textcircled{c} stand for element-wise multiplication and concatenation, respectively. The solid lines show the forward passes and the dashed lines represent the back-propagation of each loss.}
    \label{fig:proposed_pipeline}
\end{figure*}



\subsection{Conditioning Local GANs with Semantic Segmentation}
\label{sec:ss_network}

As mentioned before, earlier methods~\cite{yi2019apdrawinggan,wu2020efgan} extract local parts by rectangular bounding boxes centred around the facial landmarks to condition the local generators, which are not differential. 
This approach does not provide us flexibility to adjust the shape and size of the semantic regions and to propagate the gradients from the local GANs in the  pipeline and hence is sub-optimal. 
Here, we present a differential semantic segmentation model to extract local parts for conditioning local Generators. This network classifies each image pixel to one of the local categories.
By grouping the pixels belonging to a class any shape of a local parts can be fit, which helps us to fit semantic regions of any shape tightly as shown in Figure~\ref{fig:motivation}.

\noindent\textbf{Semantic Segmentation Network.}
For a given source RGB image $\textbf{x}_s \in \mathbb{R}^{(w\times h \times 3)}$, we feed it into $S(\textbf{x}_s; \theta_s)$,  which is a segmentation network, where $\theta_s$ are its set of learnable parameters. 
We employ state-of-the-art semantic segmentation architectures, BiSeNet~\cite{yu2018bisenet} and 
UNet~\cite{lee2020celebamask} one after another in our experiments. However, segmentation networks other than these can 
be easily applied in our pipeline.
\textbf{One of the key contributions of our method is to train the semantic segmentation and hierarchical GANs in an end-to-end manner to condition each component by the other}. To this end, inpired by \emph{soft-argmax}~\cite{chapelle2010gradient}, we apply tempered softmax on the output of the last convolutional layer of the semantic segmentation network as shown in Equation~\ref{eqn:soft_argmax}. In this Equation, $c$ is the total number of semantic categories and $\phi$ controls the temperature of the resulting probability distribution. We set the temperature $\phi$ to a positive value in order to obtain the probability of the target semantic category equal to 1 and the rest to 0.


\begin{equation}
    \begin{split}
      {M}  = \mathrm{softmax}(\phi(S(\textbf{x}_s; \theta_s))  & \\
     S:\textbf{x}_s \rightarrow S_s, \text{where}, S_s \in \mathbb{R}^{(w\times h \times c)} &
     \end{split}
\label{eqn:soft_argmax}
\end{equation}

This equation gives us $M \in \mathbb{R}^{(w \times h \times c)}$, where each channel, $M_p$, corresponds to the semantic segmentation mask for the class index $p$.
Following~\cite{wu2020efgan}, we propose to extract $4$ different local parts by using this approach. These local parts are \texttt{left eye (le)}, \texttt{right eye (re)}, \texttt{nose (n)} and \texttt{mouth (m)}. The eye regions comprise the respective eye and eyebrow classes, whereas the mouth region comprises the lower lip, upper lip and inner mouth classes.
Once the masks are obtained, for each region, we apply element-wise multiplication $\otimes$ between the input image and the semantic masks to extract the local parts, as shown in the following equation.

\begin{equation}
    \forall p \in \{le, re, n, m\}; \textbf{x}_s^{p} = {M}_{p} \otimes \textbf{x}_s
    \label{eqn:local_parts_extraction}
\end{equation}

Each extracted local part is cropped to only contain the content and resized to the size of $64 \times64$ in order to be fed into its respective local generators, while the original image $\textbf{x}_s$ is passed to the global network. We learn the parameters, $\theta_s$, of network $S$ to extract the local parts in order to maximise the performance of our translation network.

\subsection{Hierarchical GANs}
In this subsection, we present the global GAN and the local GANs guided by the semantic segmentation network. 

\noindent \textbf{Local Generators.}
For the source image $\textbf{x}_s$, conditioned on the parameters of the Semantic Segmentation network, $S(\textbf{x}_s; \theta_s)$, we infer $\textbf{x}_s^{\tt{le}}, \textbf{x}_s^{\tt{re}}, \textbf{x}_s^{\tt{n}}, \textbf{x}_s^{\tt{m}}$ as local parts for \texttt{left eye}, \texttt{right eye}, \texttt{nose}, and \texttt{mouth},
respectively as shown in Equation~\ref{eqn:local_parts_extraction}. We feed these local parts into their respective Generators.

As shown in Figure~\ref{fig:proposed_pipeline}, $\textbf{G}_{local}$:$\{\textbf{G}_{le}, \textbf{G}_{re}, \textbf{G}_{n}, \textbf{G}_{m}\}$ denotes the set of local Generators to translate \texttt{left eye}, \texttt{right eye}, \texttt{nose}, and \texttt{mouth}, respectively. 
We concatenate each local part with the same target expression label $\textbf{y}_t \in \mathbb{R}^{d}$, where $d$ is the dimension of the one-hot encoded vector representation of the target expression category. 
We create $d$ channels of dimensions matching with that of the input image by filling 0s for encoding the absence of the attributes and 1 for the presence. 
We then concatenate this vector with the image and feed it into the Encoder side of the Generator. 
The Encoder projects both the input image and the target label into lower dimension latent representations and the Decoder reconstructs the translated image as shown in Equation~\ref{eqn:local_generators}. In this Equation, $p$ represents the parts and $\textbf{x}_t^{p}$ represents the parts after translation.
\begin{equation}
\forall p \in \{le, re, m, n\}; \textbf{x}_t^{p} = \textbf{G}_{p}(\textbf{x}_s^{p}, \textbf{y}_t)
\label{eqn:local_generators}
\end{equation}
\noindent \textbf{Global Generator.} 
In addition to the local Generators, we employ a global network that serves the purpose of extracting global features of the image. As discussed before, most of the existing state-of-the-art architectures on face expression manipulation rely only on a single global generator~\cite{choi2018stargan,bhattarai2020inducing,pumarola2018ganimation}. 
These methods tend to generate artefacts and blurs especially around local areas, such as eyes, nose and mouth, where most of the manipulation takes place~\cite{wu2020efgan}. 
All of these architectures implement the generator in an encoder-decoder form. We choose to implement both the global and the local Generators by utilising the StarGAN~\cite{choi2018stargan} architecture considering both its efficiency and accuracy. However, there are no restrictions on the choice of the architecture in our framework. In Figure~\ref{fig:proposed_pipeline}, $\textbf{G}_{global}$ represents the global generator. 
Given the input image, $\textbf{x}_s$, with the target expression label, $\textbf{y}_t$, in one-hot encoded vector form, as done in local generators, we concatenate and feed them into the encoder part of the Generator. As a result, we obtain $\textbf{x}_t^{global}$, which is the image reconstructed by the decoder. 
\\

\noindent \textbf{Fusion Network.}
We employ a Fusion Network, $\textbf{G}_{fusion}$, that consists of convolutional layers to blend in the images generated by the global and the local networks. 
This network takes the concatenation of the output of the local generators and the global generator as input and gives us the final translated image $\textbf{x}_t$.
Since we have passed only the local parts to the local generators, we stitch them together by placing each local region centred to its original location in $\textbf{x}_s$. By doing so we obtain $\textbf{x}_{t}^{local}$ that contains all local regions and matches the dimensions of the global network. 
Afterwards, we concatenate $\textbf{x}_{t}^{local}$ and $\textbf{x}_{t}^{global}$ and pass it to $\textbf{G}_{fusion}$, which consists of two convolution blocks and four residual blocks followed by a convolution layer. 
This network generates the final image $\textbf{x}_t$, which is passed to the discriminator $\textbf{D}_{fusion}$.
\begin{equation}
\begin{split}
\textbf{x}_f = \textbf{x}_{t}^{global} \oplus \textbf{x}_{t}^{local} \\
\textbf{x}_t = \textbf{G}_f(\textbf{x}_f) &
\end{split}
\end{equation}

\noindent \textbf{Discriminator.} Each of the generators are equipped with their respective discriminators, and the respective generators and discriminators compete against each other in an adversarial manner to learn their parameters.  \\
\noindent\textit{Adversarial Loss.} The overall adversarial loss is the sum of the adversarial losses for the global network, fusion network and the local networks. 
In Equation~\ref{eqn:adv_loss}, $p$ represents the local parts, the global network and the fusion network.

\begin{equation}
\begin{split}
    \mathcal{L}^p_{adv}= &\mathbb{E}_{\textbf{x}_s^{p}} [\log(\textbf{D}_p({\textbf{x}_s^{p}}))] \\ 
    &+ \mathbb{E}_{\textbf{x}_s^{p},\textbf{y}_t} [\log(1 - \textbf{D}_p(\textbf{G}_p(\textbf{x}_s^{p}, \textbf{y}_t))] \\
    & \forall p \in \{global, fusion, le, re, m, n\}
\label{eqn:adv_loss}
\end{split}
\end{equation}


\noindent\textit{Classification Loss.} Besides being realistic, in expression translation the synthesised image, $\textbf{x}_t$, should also bear the given target expression, $\textbf{y}_t$. Hence, we add an auxiliary classifier, $\textbf{D}_p^{cls}$, on top of $\textbf{D}_p$, which guides $\textbf{G}_p$ to generate images that will be confidently classified as $\textbf{y}_t$. This classification loss is defined as follows:

\begin{equation}
    \begin{split}
    \mathcal{L}^p_{cls}=\mathbb{E}_{\textbf{x}_s^{p}, \textbf{y}_t} [\log(\textbf{D}_{p}^{cls}(\textbf{y}_t|\textbf{G}_p(\textbf{x}_s^{p}, \textbf{y}_t)))] \\
     \forall p \in \{global, fusion, le, re, m, n\} &
    \label{eqn:cls_loss}
    \end{split}
\end{equation}





\noindent\textit{Reconstruction Loss.} In order to preserve the identity while translating the facial expression, we employ the cycle loss \cite{zhu2017cyclegan}. This loss is defined as

\begin{equation}
    \mathcal{L}^p_{rec}=\mathbb{E}_{\textbf{x}_s^{p}}[\|\textbf{x}_s^{p}-\hat{\textbf{x}}_{s}^{p}\|_1] 
\label{eqn:res_loss}
\end{equation}

where the reconstructed image, $\hat{\textbf{x}}_{s}^{p}$, is obtained by $\textbf{G}_p(\textbf{x}_t^p, \textbf{y}_s)$. Here, $\textbf{x}_t^{p}$ is the generated image and $\textbf{y}_s$ is the  expression of the input image, $\textbf{x}_s$. 

\noindent\textbf{Overall Training Objective.} The overall objective loss is summarised as:

\begin{equation}
\begin{split}
    \mathcal{L} = &\sum_{p} \lambda_p (\lambda_{adv}{\mathcal{L}^p_{adv}} + {\lambda_{cls}}{\mathcal{L}^p_{cls}} + {\lambda_{rec}}{\mathcal{L}^p_{rec}}) \\
    &\forall p \in \big\{global, fusion, local=\{le, re, m, n\}\big\}
\label{eqn:obj_loss}
\end{split}
\end{equation}

\subsection{Conditioning Semantic Segmentation with GANs}
\label{methodology:cond_ss}
We explained the process of updating the parameters of every learnable component of the hierarchical GANs.
Unlike previous methods~\cite{wu2020efgan,yi2019apdrawinggan}, which rely on pre-defined static local parts, we also learn the extraction of the local parts. 
Please recall that our framework is end-to-end differentiable. Hence, while propagating the gradients we do not only propagate to the Generators and the Discriminators but also to the semantic segmentation network. The set of the parameters of the semantic segmentation network $\theta_s$ are updated with the gradients computed with respect to the loss incurred by the local networks and the fusion network as shown in Equation~\ref{eqn:condn_sem_seg}. To the best of our knowledge, this is the first framework to condition semantic segmentation networks and hierarchical GANs in both directions, training the network in an end-to-end manner.

\begin{equation}
\theta_s = \theta_s - \lambda_p\bigg(\frac{\partial \mathcal{L}_{fusion}}{\partial\theta_s} + \frac{\partial \mathcal{L}_{local}}{\partial\theta_s}  \bigg)
\label{eqn:condn_sem_seg}
\end{equation}


%% file: experiments.tex
\section{Experiments}

\begin{table*}[t]
    \centering
    \footnotesize
    \resizebox{0.85\textwidth}{!}{
    \begin{tabular}{c|c|c|c|c|c|c|c}
     \toprule[1pt] 
     \textbf{Translation Dataset} & \textbf{Method} & \textbf{Year} &\textbf{EEA (\%) $\uparrow$} & \textbf{FID$\downarrow$} & \textbf{ID Loss$\downarrow$} & \textbf{Segmentation Dataset} & \textbf{mIOU} $\uparrow$\\
     \midrule[0.5pt]
     RaFD & StarGAN~\cite{choi2018stargan} & CVPR'18 & 90.2  & 14 & 0.23  &- & - \\
     RaFD &Ganimation \cite{pumarola2018ganimation} & ECCV'18& 82.4 & 17.16 & n/a & - & - \\
     RaFD &EF-GAN \cite{wu2020efgan} &CVPR'20 & 93.2  & 13.64 &  0.21 & - & - \\
    
    RaFD &Ours w/o end-to-end  & -& \textbf{94.1 (+3.9)} & 13.46 & 0.2 & CelebAMask-HQ  & 72.84\\
    RaFD &Ours w/ end-to-end  & -& \textbf{95.3 (+5.1)} & \textbf{13.09} & 0.19 & CelebAMask-HQ & \textbf{76.22}\\
     \midrule[0.5pt]
     AffectNet & StarGAN~\cite{choi2018stargan} & CVPR'18 &82.1  & 16.33 & 0.36 & - & - \\

     Affectnet & Ganimation~\cite{pumarola2018ganimation} & ECCV'18& 74.3 & 18.86 &  n/a & - & - \\
     Affectnet & EF-GAN~\cite{wu2020efgan} &CVPR'20 & 84.4 & 13.84 &  0.28 & - & - \\
    Affectnet & Ours w/o end-to-end & - &\textbf{89.7 (+7.6)} & 12.03 & 0.24 & CelebAMask-HQ & 72.84 \\
    Affectnet & Ours w/ end-to-end & - &\textbf{90.5 (+8.4)} & \textbf{11.8} & 0.24 & CelebAMask-HQ & \textbf{77.60} \\

     \bottomrule[1pt] 
    \end{tabular}
    }
    \caption{\textbf{Performance Comparison on translation datasets, RaFD and AffectNet, and segmentation dataset CelebAMask-HQ.} Our method outperforms compared methods in EEA, FID, ID loss while also improving the semantic segmentation performance without any need of additional segmentation annotations.}
    \label{tab:exp_compare_affectnet}
\end{table*}

\noindent\textbf{Datasets.} We evaluate the expression manipulation performance of our method on AffectNet \cite{mollahosseini2019affectnet} and Radbound Faces Dataset (RaFD) \cite{langner2010rafd}. 
AffectNet contains $450K$ manually annotated images, which we utilised for our experiments.
RaFD consists of $8,040$ images of 67 models, with 5 different angles and 3 gazes. In parallel with EF-GAN\cite{wu2020efgan}, as profile images do not contain all local parts, for RaFD we use only 90$^{\circ}$ angled faces, leading to 1,608 images. We pre-process the images by aligning them based on their landmarks and crop them to the size $256\times256$. We pre-train the semantic segmentation network on CelebAMask-HQ training set and report the performance on its test set. There are $30K$ facial images with fine-grained mask annotations. 

\noindent\textbf{Implementation.} We implement our method on PyTorch \cite{paszke2017pytorch}. We trained our model with a batch size of $8$ for $200K$ epochs. We apply Adam optimiser \cite{kingma2014adam} ($\beta_1 = 0.5, \beta_2 = 0.999$) with a learning rate of $0.0001$. 
We perform our experiments on a workstation with 32G memory, NVIDIA GTX1080.
We implement local and global GANs with StarGAN \cite{choi2018stargan} architecture. 
With these settings, it takes approximately 30 hours to train the whole proposed network on AffectNet, whereas training the baseline, e.g StarGAN, takes around 12 hours.

\noindent\textbf{Evaluation Metrics.} 
We calculate the \emph{Frechet Inception Distance (FID)} \cite{heusel2017fid}, \emph{Expression Editing Accuracy (EEA)} and \emph{Identity (ID) loss}. FID is a commonly used metric to evaluate the performance of GANs by assessing the quality and diversity of generated images. 
EEA is another important metric to check the fidelity of the translated expression.
Similar to~\cite{choi2018stargan}, we train VGG-16~\cite{simonyan2015vgg} as an expression classifier on the real train set and report the performance on the images translated from the real-test set. 
In addition, we compute the cosine distance of ArcFace~\cite{deng2018arcface} embeddings
to measure the ID loss for the synthesised images. This metric gives us an idea about how well an architecture is able to disentangle identity from expression. 
We measure the performance of the semantic segmentation network by the~\emph{mean Intersection-over-Union (mIoU)} metric. Finally, we present extensive qualitative comparisons for both synthesis and segmentation. We suggest to the reader to refer to the supplementary material for additional results.

\noindent\textbf{Compared Methods.} 
We compared our method to three different architectures: StarGAN~\cite{choi2018stargan}, GANimation~\cite{pumarola2018ganimation} and EF-GAN~\cite{wu2020efgan}. StarGAN and GANimation are two important and state-of-the-art methods for expression manipulation, whereas EF-GAN is the state-of-the-art hierarchical GAN for expression translation. For semantic segmentation, we took BiSeNet~\cite{yu2018bisenet} and UNet~\cite{lee2020celebamask} trained on CelebAMask-HQ as baselines. We compare the performance of these networks before and after fine-tuning them by our pipeline.
Please note that we employ a pre-trained segmentation network to extract local parts, and expression translation is carried out on AffectNet and RaFD, which do not have semantic segmentation annotations.

\subsection{Quantitative Evaluations}

\begin{table}[t]
    \centering
    \footnotesize
    \resizebox{0.83\linewidth}{!}{
    \begin{tabular}{c|c|c|c}
     \toprule[1pt] 
     \textbf{Network} &\textbf{Method} &  \textbf{mIoU (\%) $\uparrow$} & \textbf{Translation Dataset} \\

    \midrule[0.5pt]
    BiSeNet\cite{yu2018bisenet} & Baseline & 72.84 & -\\ 

     & Ours & \textbf{77.60 (+4.76)} & AffectNet \\ 
     
     & Ours & \textbf{76.22 (+3.38)} & RaFD \\ 
     
     \bottomrule[1pt] 
     
    UNet\cite{lee2020celebamask} & Baseline & 71.66 & -\\ 

     & Ours & \textbf{73.10 (+1.44)} & AffectNet \\
     & Ours & \textbf{73.00 (+1.34)} & RaFD \\
     
     \bottomrule[1pt] 
    \end{tabular}
    }
    \caption{\textbf{Performance of the semantic segmentation network on CelebAMask-HQ.} 
    }
    \label{tab:exp_compare_parser}
\end{table}

\begin{figure*}[t]
\centering
\includegraphics[width=0.42\linewidth]{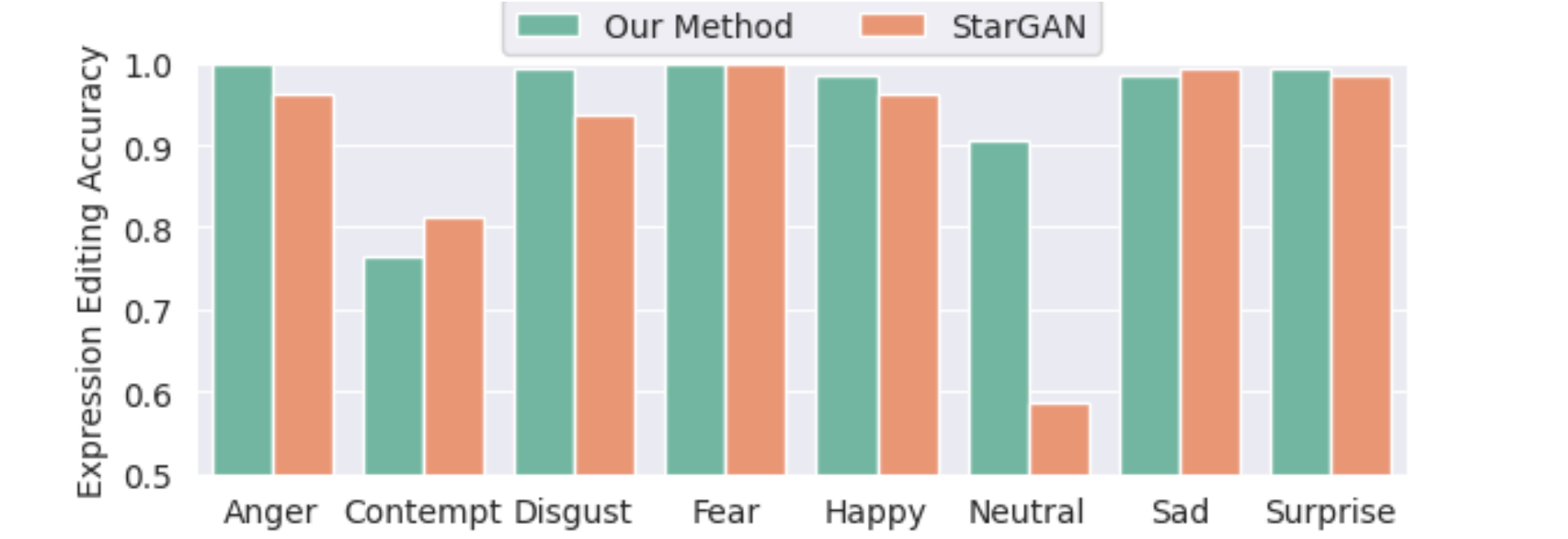}
\includegraphics[width=0.4\linewidth]{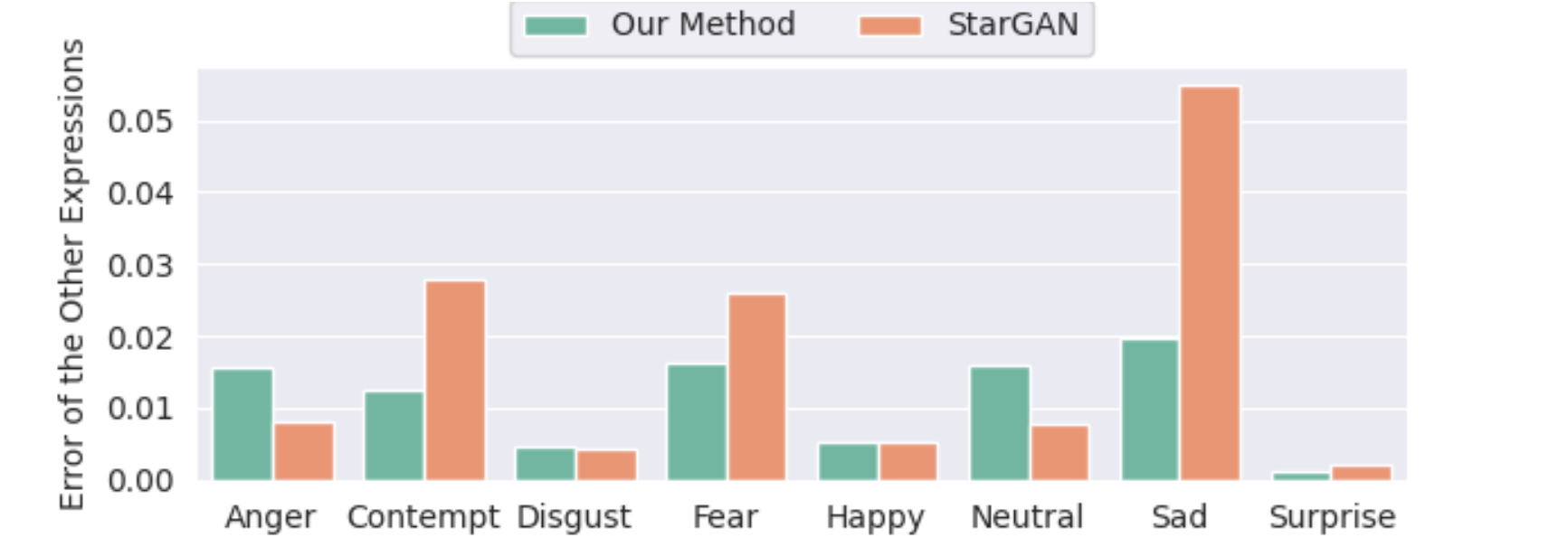}
\\ (a) Expression Editing Accuracy
\hspace{2 cm}
(b) Expression Preservation Error
\caption{\textbf{Expression editing accuracy (a) and expression preservation error (b).} Comparison of our method to StarGAN on RaFD.}
\label{fig:rafd_expression_editing_preserving}
\end{figure*} 

\noindent\textbf{Expression Editing Accuracy (EEA).} 
Table~\ref{tab:exp_compare_affectnet} (fourth column) summarises the EEA on both RaFD and AffectNet. 
Our method has two variants: \emph{w/o end-to-end} depicts no propagation of gradients beyond hierarchical GANs on the backward pass, while in case of \emph{w/ end-to-end} the gradients are back-propagated till the semantic segmentation network.
For images translated by our method on RaFD, the EEA is $95.3\%$, which corresponds to a $5.1\%$ improvement over StarGAN, i.e our global network, and a $2.1\%$ improvement over EF-GAN, the current state-of-the-art method. On AffectNet, our method outperforms all compared methods with an improvement over StarGAN by $8.4\%$ and EF-GAN by $6.1\%$.
These results highlight the advantage of hierarchical networks over a single global network as well as using semantic segments over rectangular regions as a method to extract local regions.
\emph{End-to-end} training improves the performance of both expression editing and semantic segmentation (last column) demonstrating the symbiotic nature of these two tasks.    
Figure~\ref{fig:rafd_expression_editing_preserving} shows the expression-wise performance comparison (a) and the preservation error of other expressions (b). Our method is more robust compared to the baseline and produces a lower error rate for most classes, which shows that it translates facial expressions with higher accuracy by modifying only the parts related to the target expression.

\begin{figure}
    \centering
    \includegraphics[trim= 0.5cm 0.2cm 0cm 0.3cm, clip, width=0.65\linewidth]{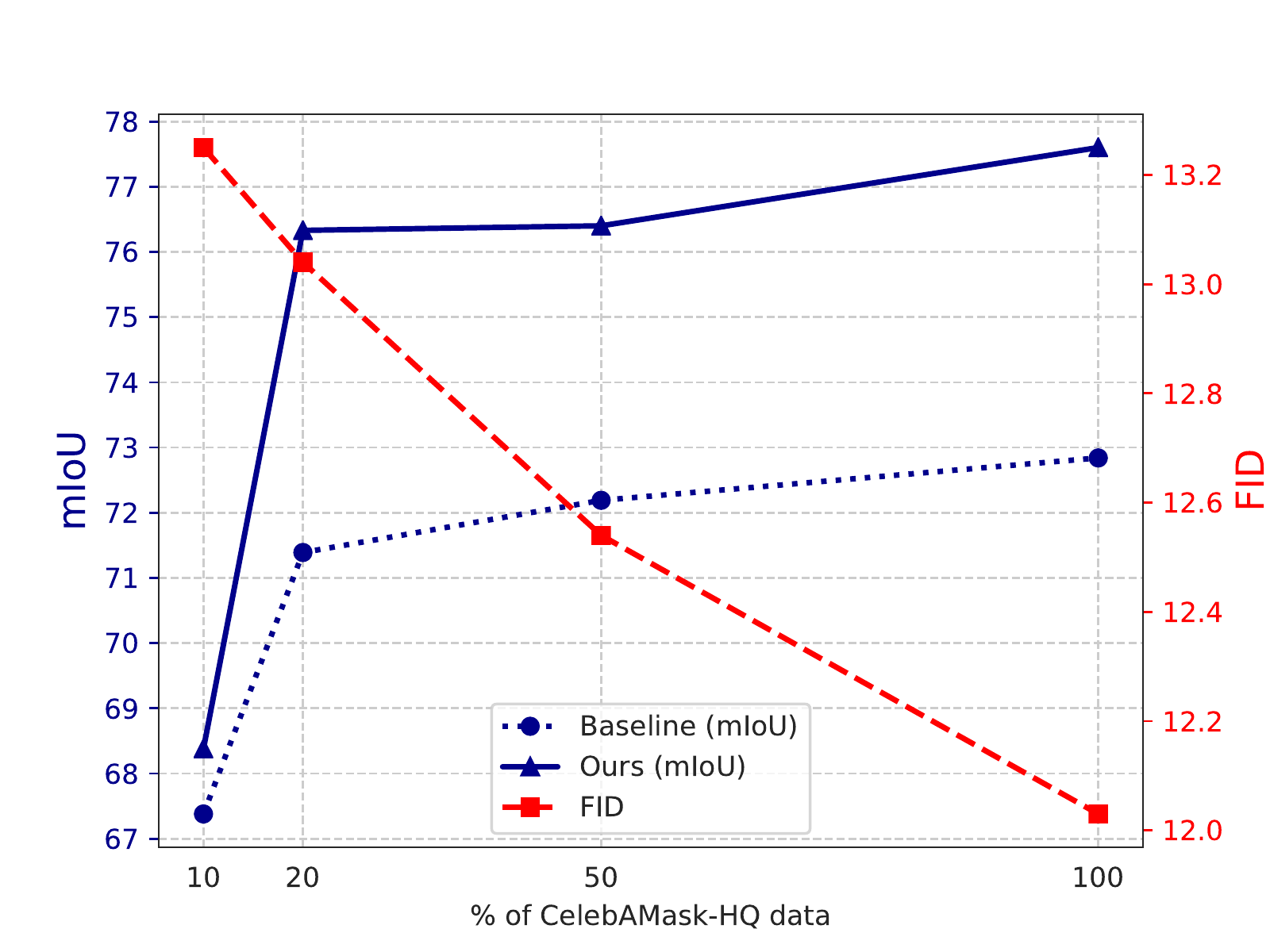}
    \caption{\textbf{mIOU and FID with different sizes of training samples of CelebAMask-HQ.
    } } 
    \label{fig:subset_parser}
\end{figure}

\noindent\textbf{FID.} To further evaluate the synthesised images, we calculate the FID for both datasets. On both AffectNet and RaFD, we observe an FID score (see Figure~\ref{tab:exp_compare_affectnet}, fifth column) lower than all compared methods. Lower FID scores indicate better quality and diversity in synthesised images.
\noindent\textbf{ID Loss.} We assess how well the ID is preserved during translation by computing the cosine distance of ArcFace embeddings~\cite{deng2018arcface} between the synthetic images and their corresponding real images. The results in Table \ref{tab:exp_compare_affectnet} show that our method accomplishes to preserve the ID better while synthesising images of higher classification accuracy.   \\
\noindent\textbf{mIoU.} Table~\ref{tab:exp_compare_parser} shows the performance comparison of the semantic segmentation network before and after the end-to-end training of our model. The baseline refers to the segmentation model that is pre-trained on the training set of CelebAMask-HQ, which yields an mIoU rate of $72.84\%$. The network parameters are updated during the end-to-end training of our method on the given translation datasets without any ground truth segmentation annotation. mIoU of the semantic segmentation network improves to $77.6\%$ and $76.22\%$ when updated on AffectNet and RaFD, respectively. 
By replacing the backbone with UNet~\cite{lee2020celebamask} we observe a similar trend. 
As presented in Figure~\ref{fig:subset_parser}, we took different proportions of CelebAMask-HQ training set (10\%, 20\%, 50\%) to pre-train the semantic segmentation network. We used these models to separately initilise and train our pipeline end-to-end.
Two important observations to note are that (i) only $20\%$ of the segmentation annotations are required by our method to outperform the baseline trained on the whole dataset, and (ii) the quality of synthetic data (see FID on right axis) is correlated with the performance of the segmentation network.  
This shows the capability of our method to utilise annotations from heterogeneous datasets for improving the performance of a downstream task. Our method would be crucial for tasks like segmentation in which annotations need to be done on pixel level.

\begin{figure}
    \centering
    \includegraphics[trim= 0.1cm 0.1cm 0.1cm 0.1cm, clip,width=0.3\textwidth]{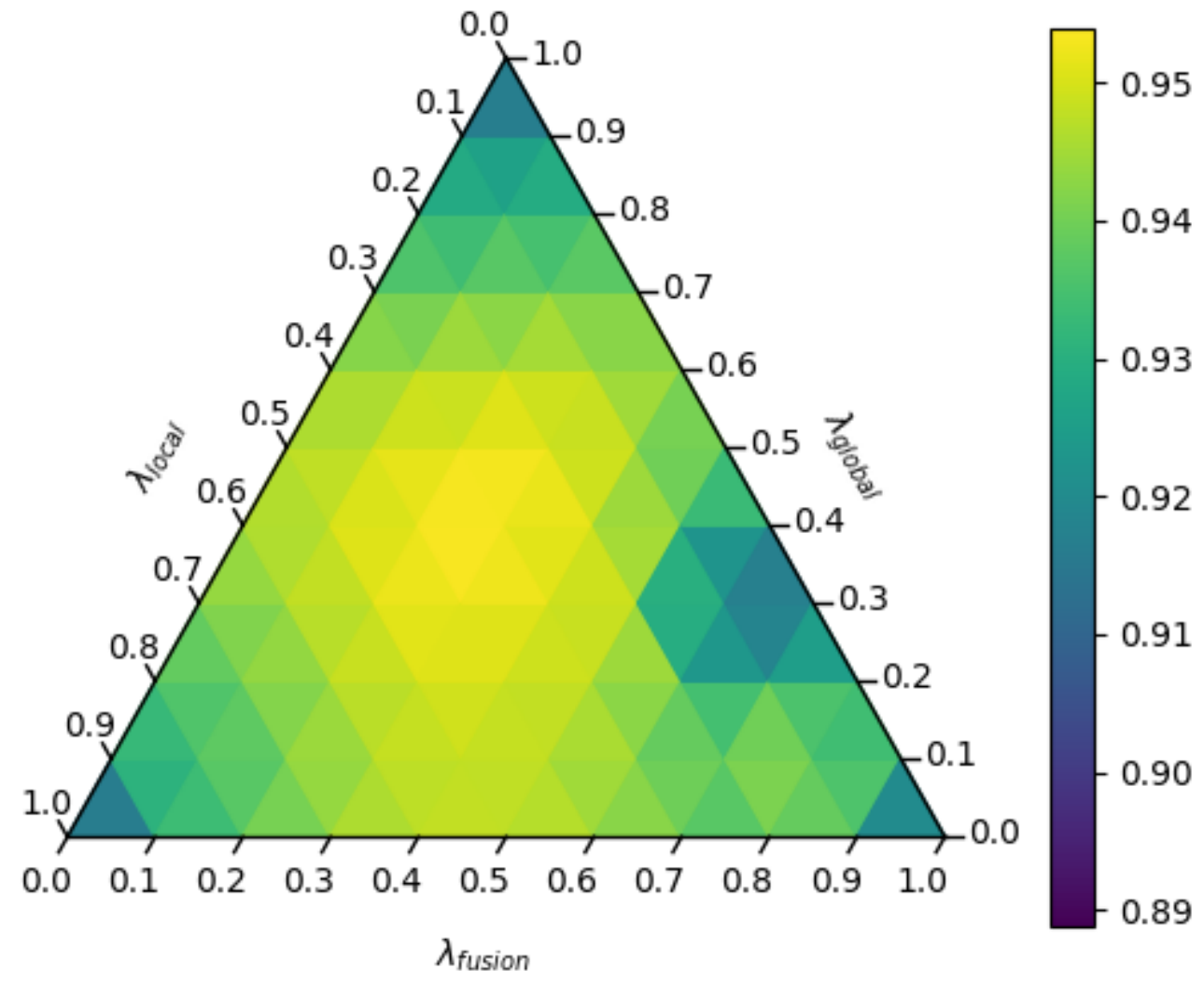}
    \caption{\textbf{Hyper-parameter tuning on RaFD.} 
    (See Eqn.~\ref{eqn:obj_loss})}
    
    \label{fig:finetuning}
\end{figure}

\noindent\textbf{Hyper-parameter Study.} We set different values for the weights of the losses incurred by each network in Eqn.~\ref{eqn:obj_loss}. Figure \ref{fig:finetuning} visualises this study in the form of a ternary heatmap that shows the classification accuracy, where the axes represent the weights for the global, local and fusion networks. 
We observe that incorporating local features improves over using only global loss, and an accuracy rate up to $95.3\%$ is reached when the weights are fine-tuned.  


\begin{figure*}
    \centering
    \includegraphics[width=0.7\linewidth]{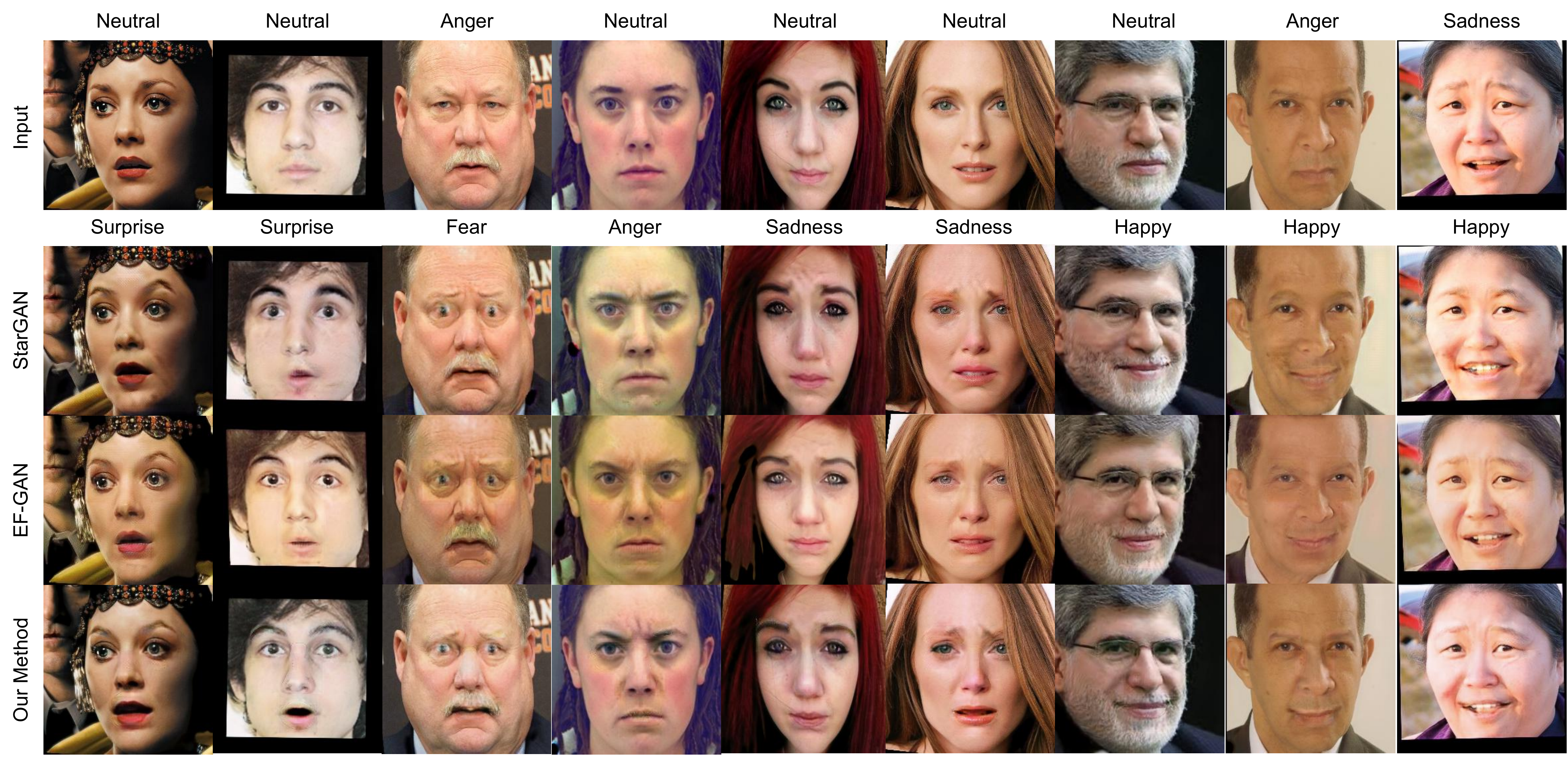}
    \caption{\textbf{Qualitative comparisons of images translated by StarGAN, EF-GAN and our method on AffectNet.} Images synthesised by our method are more realistic, bear the target expression and preserve the identity better than compared methods, e.g StarGAN fails to preserve identity related features of the eyebrows in the second and third samples, both StarGAN and EF-GAN fail to translate the mouth in the second sample.}
    \label{fig:qualitative_affectnet}
\end{figure*}

\noindent\textbf{Ablation Study.} We examined the effect of each local region by excluding them during training as summarised in Table \ref{tab:exp_localregions}. We observe that in comparison to other local parts, the nose region has the smallest effect on the expression translation performance whilst the eyes have the largest. This is expected and in accordance with qualitative results as the least change happens in the nose area, which bears less information regarding the expression.

\subsection{Qualitative Evaluations}

\begin{figure*}
    \centering
    \includegraphics[ clip,width=0.85\linewidth]{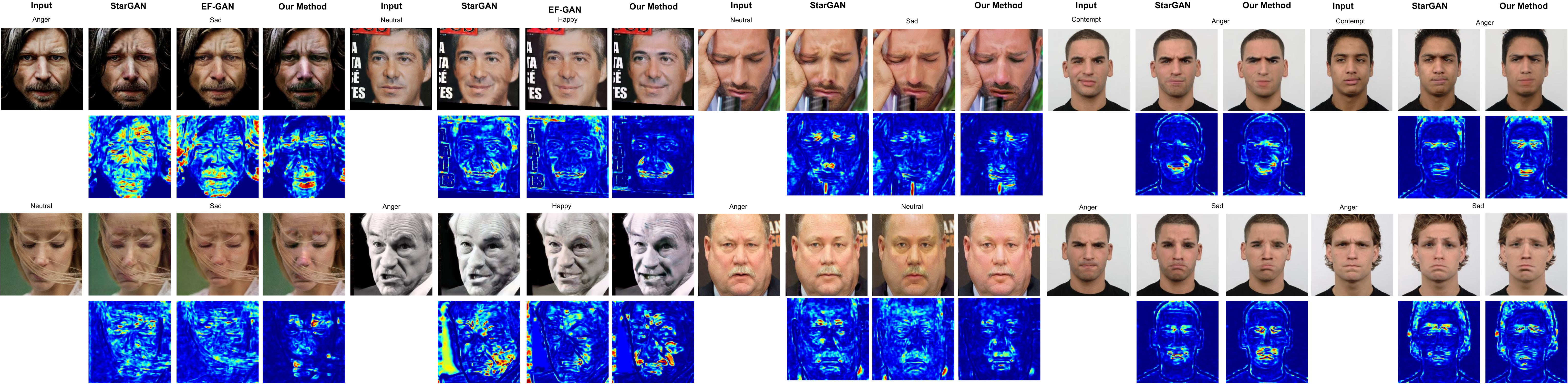}
    \caption{\textbf{Comparisons of heatmaps showing the attention and artefacts on translated images of AffectNet (first three vertical blocks) and RaFD (last two vertical blocks).} The heatmaps below each generated image shows their pixel difference from the input image. Our method focuses on expression-related regions while compared methods cause unwanted changes on identity-related areas.} 
    \label{fig:attention_map}
\end{figure*}

\noindent\textbf{Expression Manipulation.} Figure \ref{fig:qualitative_affectnet} shows a comparison of the images synthesised by StarGAN, EF-GAN and our method on AffectNet. As seen, our method synthesises more natural-looking faces that bear the target attribute while maintaining the identity. In many cases, we observe that StarGAN fails to capture local details and preserve identity related features. For instance, on the first sample, we can see artefacts on the chin and the cheeks, while on the second sample the mouth is unclear, the nose and skin contain artefacts and the eyebrows become bushy. Similarly, on the third sample the original texture of the eyebrow is not preserved while modifying the expression. 
Although EF-GAN is better than StarGAN at capturing such local details, it still fails to synthesise local details realistically despite the usage of local networks. Moreover, as the local parts in EF-GAN are extracted in rectangular shapes with predefined sizes, we observe some texture inconsistency and traces of the rectangular shape in some samples. For example, the eye areas on the second sample from right and the third sample show a difference on the skin texture. This is caused by the inaccurate local parts fed into the local generators. These results illustrate that, while employing local networks improves over using only a global network, it is crucial to extract the local regions accurately and precisely.

In Figure \ref{fig:attention_map}, we show further results on AffectNet and RaFD along with heatmaps illustrating pixel differences between the input and the synthesised image. 
We observe that with our method the manipulation occurs mostly on the regions related to the expression, whereas both StarGAN and EF-GAN manipulate also unrelated areas. Especially in cases where the face is not frontal or there are occlusions, both methods lead to results that contain artefacts. For instance, on the bottom left, the attention of the compared methods is mostly on the occluding hair. Similarly, in the second top sample, 
StarGAN and EF-GAN cause unwanted modifications on the hair and background. Our method, on the other hand, manages to correctly modify the expression while preserving identity related details. 

\begin{table}[t]
    \centering
    \footnotesize
    \resizebox{0.7\linewidth}{!}{
    \begin{tabular}{c|c|c}
     \toprule[1pt] 
     Regions Used &  Accuracy (\%) $\uparrow$ & FID $\downarrow$\\

    \midrule[0.5pt]
    \texttt{full} & 90.5 & 11.8\\
    \texttt{full - mouth} & 86.9 & 13.52\\ 
    \texttt{full - nose} & 88.6 & 12.84   \\
    \texttt{full - eyes} & 85.2 & 15.5 \\

     \bottomrule[1pt] 
    \end{tabular}
    }
    \caption{\textbf{Ablation study for local regions on AffectNet.} 
    Here, \texttt{full} denotes the mask and \texttt{-} indicates the exclusion of the local part.
    }
    \label{tab:exp_localregions}
\end{table}

\begin{figure}
    \centering
    \includegraphics[trim= 0cm 0cm 0cm 0cm, clip,width=\linewidth]{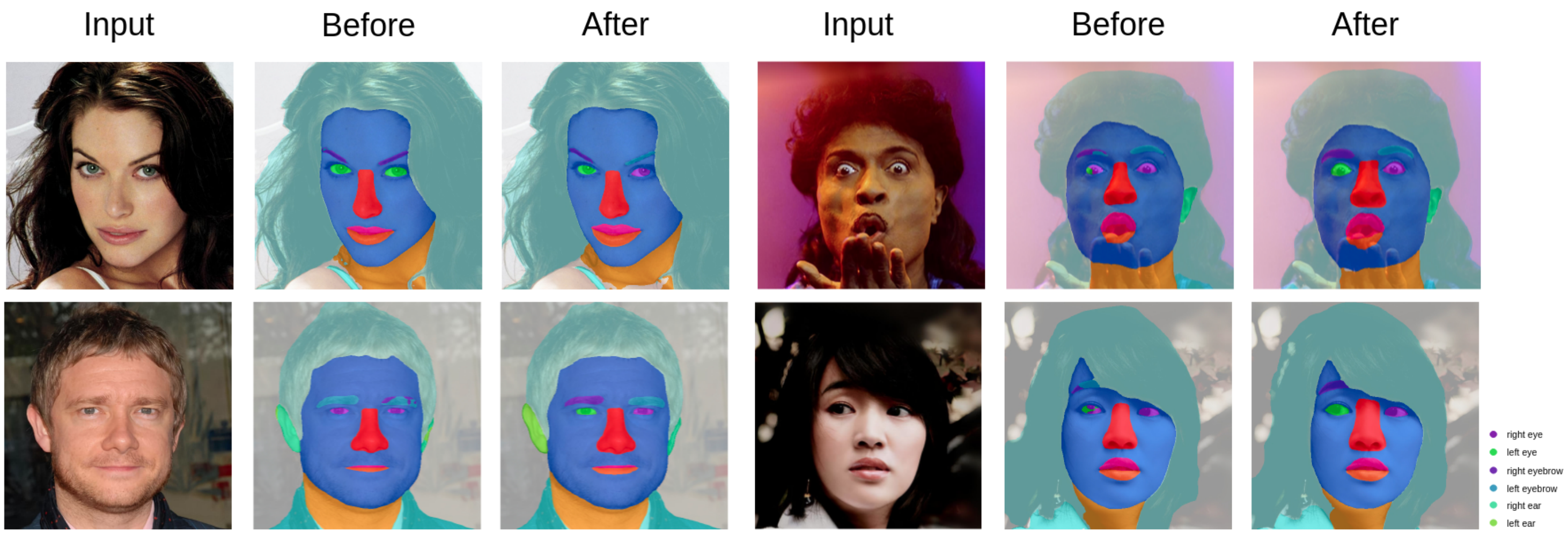}
    \caption{\textbf{Semantic segments on CelebAMask-HQ before and after the end-to-end training off our method.} The end-to-end training of our model on AffectNet, without any semantic segmentation annotation improves the performance of the semantic segmentation network. (Zoom in for better view)} 
    \label{fig:qualitative_semantic_segmentation}
\end{figure}

\noindent\textbf{Semantic Segmentation.} Figure \ref{fig:qualitative_semantic_segmentation} shows a comparison of the semantic segmentation maps obtained before and after updating the semantic segmentation network with our method.
Before updating this network, 
it is pre-trained on the training set of CelebAMask-HQ. The parameters of this pre-trained model are updated during our end-to-end training, without using any segmentation annotation. 
In parallel with the results presented on Table \ref{tab:exp_compare_parser}, Figure \ref{fig:qualitative_semantic_segmentation} shows that updating the network parameters with our method improves the segmentation results. We observe that the pixels are classified more accurately to the relevant semantic regions, particularly the eyes and eyebrows.  

%% file: conclusion.tex
\section{Conclusion}

In this work, we propose a unified framework of hierarchical GANs and semantic segmentation for expression translation, where both models benefit from each other. On forward pass, the local GANs that focus on local details are conditioned by a semantic segmentation network, which is optimised to extract local regions of the face. Unlike previous methods, this enables the model to adjust the size and shape of the local regions accordingly. 
On backward pass, we condition the semantic segmentation network by propagating gradients from the hierarchical GANs constructing an end-to-end differentiable architecture. Expression translation 
and semantic segmentation experiments 
demonstrate the effectiveness of our method over the state-of-the-art.

\noindent\textbf{Ethics statement.}  We note that, like other image generation techniques, our method is also vulnerable to being used for synthesising fake content without consent and deceiving the public. To prevent such potential misuses, methods that are proposed for detecting fake content or encoding artificial fingerprints should be utilised.

%% file: limitations.tex
\noindent \textbf{Limitations.}
The addition of local networks increases the overall complexity of the model in comparison to the baseline as reported. 
In addition, the semantic segmentation network used by our method needs to be pretrained with segmentation annotation. 

\subsection*{Acknowledgements}

R. Bodur is funded by the Turkish Ministry of National Education. This work is partly supported by EPSRC Programme Grant ‘FACER2VM’(EP/N007743/1).

%% file: supplementary.tex
\section*{Appendix}
\beginsupplement

In this supplementary material we present more qualitative experimental results in order to provide a 
comparison of our method to the state-of-the-art.

\section*{Additional Qualitative Results}
\begin{figure*}
    \centering
    \includegraphics[width=\linewidth]{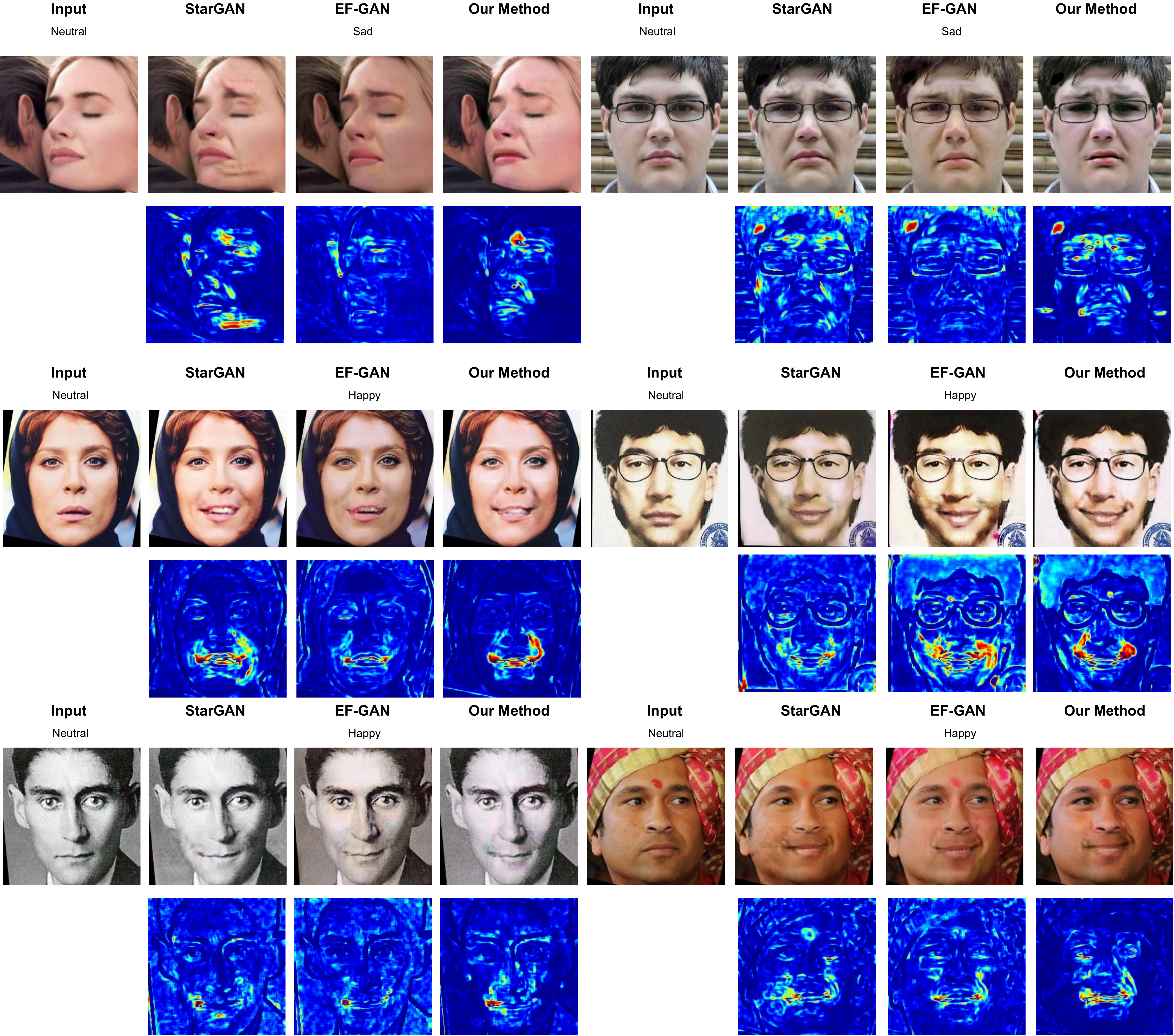}
    \caption{\textbf{Comparisons of heatmaps showing the attention and artefacts on translated images of AffectNet.}}
    \label{fig:qualitative_heatmap}
\end{figure*}

\begin{figure*}
    \centering
    \includegraphics[width=\linewidth]{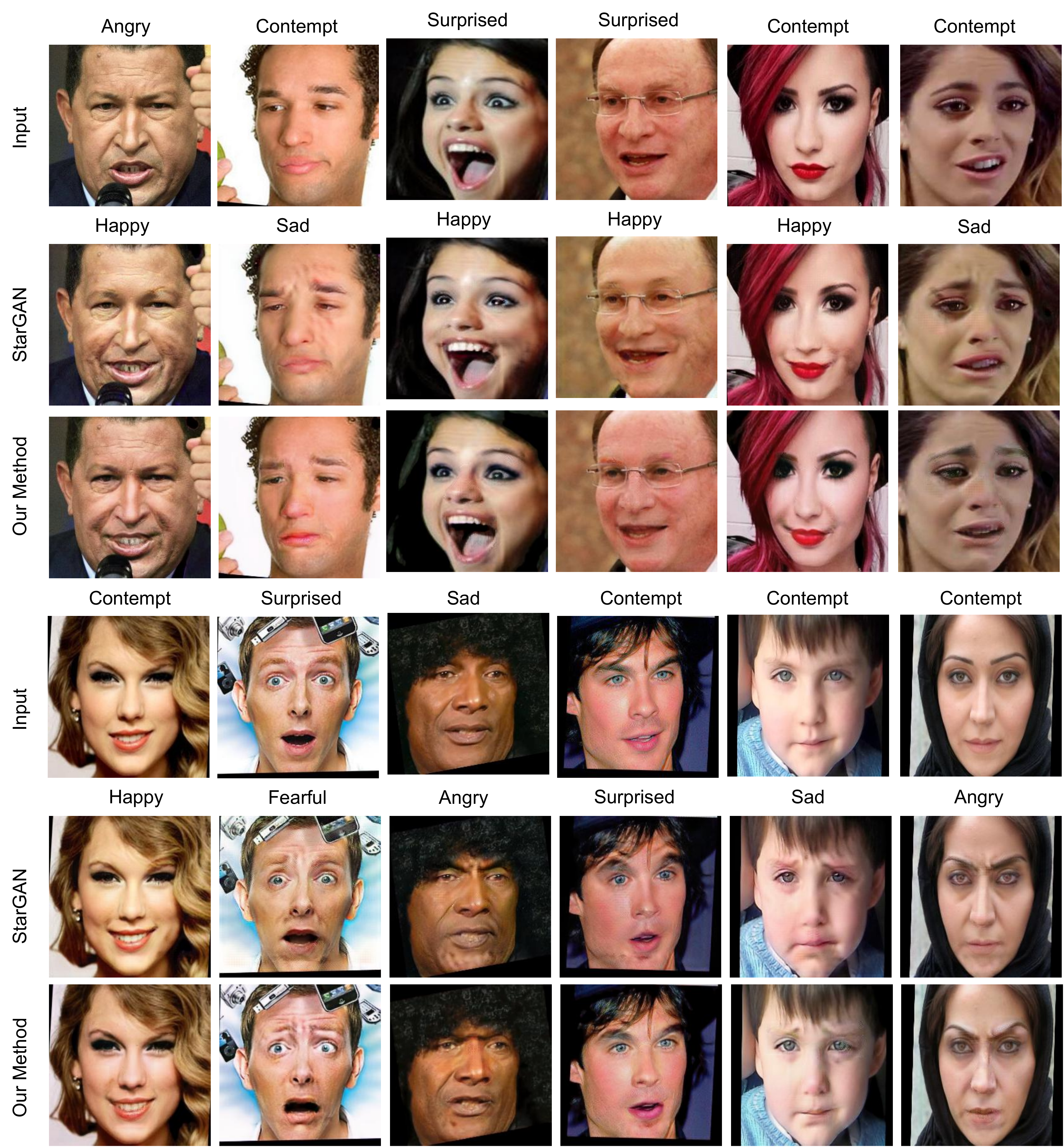}
    \caption{\textbf{Qualitative comparisons of images translated by StarGAN and our method on AffectNet.}}
    \label{fig:qualitative_affectnet_2}
\end{figure*}

\noindent\textbf{Expression Manipulation.} Figure \ref{fig:qualitative_affectnet_2} provides a qualitative comparison of images synthesised by our method and StarGAN. We observe that our method successfully translates from one expression to the other while preserving identity related features and preventing artefacts that mostly occur on local regions.  For instance, when translating from contempt to anger on the last sample, with StarGAN the eye colour changes, the eyebrows become thicker and contain artefacts and blurs.  
In Figure \ref{fig:qualitative_heatmap}, we further compare StarGAN and EF-GAN to our method by providing heatmaps that illustrate the pixel differences between the input image and the synthesised image. For example, on the first sample the image synthesised by StarGAN contains artefacts on the chin, cheeks and the eyebrows and EF-GAN results in texture inconsistency in local regions whereas our method translates the image with little or no artefacts.

\begin{figure*}
    \centering
    \includegraphics[width=0.95\linewidth]{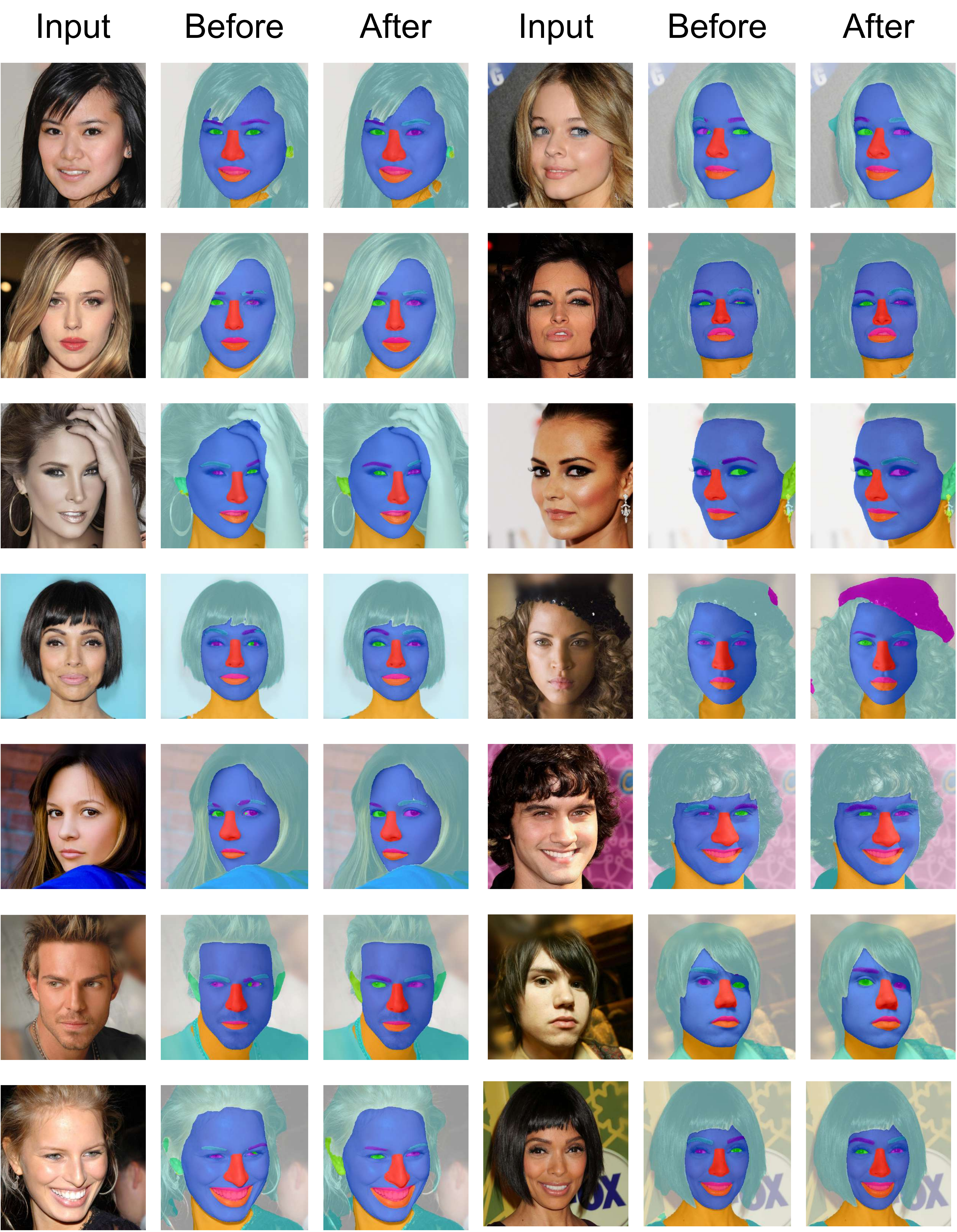}
    \caption{\textbf{Semantic segments on CelebAMask-HQ before and after the end-to-end training of our method.}}
    \label{fig:qualitative_ss_2}
\end{figure*}

\noindent\textbf{Semantic Segmentation.} Figure \ref{fig:qualitative_ss_2} shows a visual comparison of semantic segmentation maps obtained before and after the end-to-end training of our method. We observe that our method improves the overall performance of the semantic segmentation network. In particular, pixels belonging to eyes and eyebrows are assigned more accurately to the correct class. 

\noindent\textbf{Semantic Segmentation Network Architecture.} In order to evaluate the effect of different semantic segmentation architectures on the performance of our method, we experimented with two different architectures, BiSeNet and UNet, as summarised in Table~\ref{tab:exp_ss_arch}. We observe that regardless of the choice of architecture, our method outperforms the baseline, and the end-to-end training improves the semantic segmentation performance.

\begin{table*}[t]
    \centering
    \footnotesize
    \resizebox{0.9\linewidth}{!}{
    \begin{tabular}{c|c|c|c|c}
     \toprule[1pt] 
     Method & SS Architecture &  Accuracy (\%) $\uparrow$ & FID $\downarrow$ & mIoU\\ 

    \midrule[0.5pt]
    StarGAN & - & 82.1  & 16.33 & - \\
    \hdashline
    Ours w/o end-to-end & BiSeNet &\textbf{89.7 (+7.6)} & 12.03 (-4.3) & 72.84  \\
    Ours w/ end-to-end & BiSeNet &\textbf{90.5 (+8.4)} & \textbf{11.8} (-4.53) & \textbf{77.60} (+4.76) \\
    \hdashline
    Ours w/o end-to-end & UNet &\textbf{89 (+6.9)} & 12.3 (-4.03) & 71.66   \\
    Ours w/ end-to-end & UNet &\textbf{89.4 (+7.3)} &  \textbf{12.09 (-4.24)} & \textbf{73.1(+1.44)}\\
    \bottomrule[1pt] 
    \end{tabular}
    }
    \caption{\textbf{Performance of our method with different semantic segmentation network architectures on AffectNet.} 
    }
    \label{tab:exp_ss_arch}
\end{table*}

\begin{figure*}
    \centering
    \includegraphics[width=0.95\linewidth]{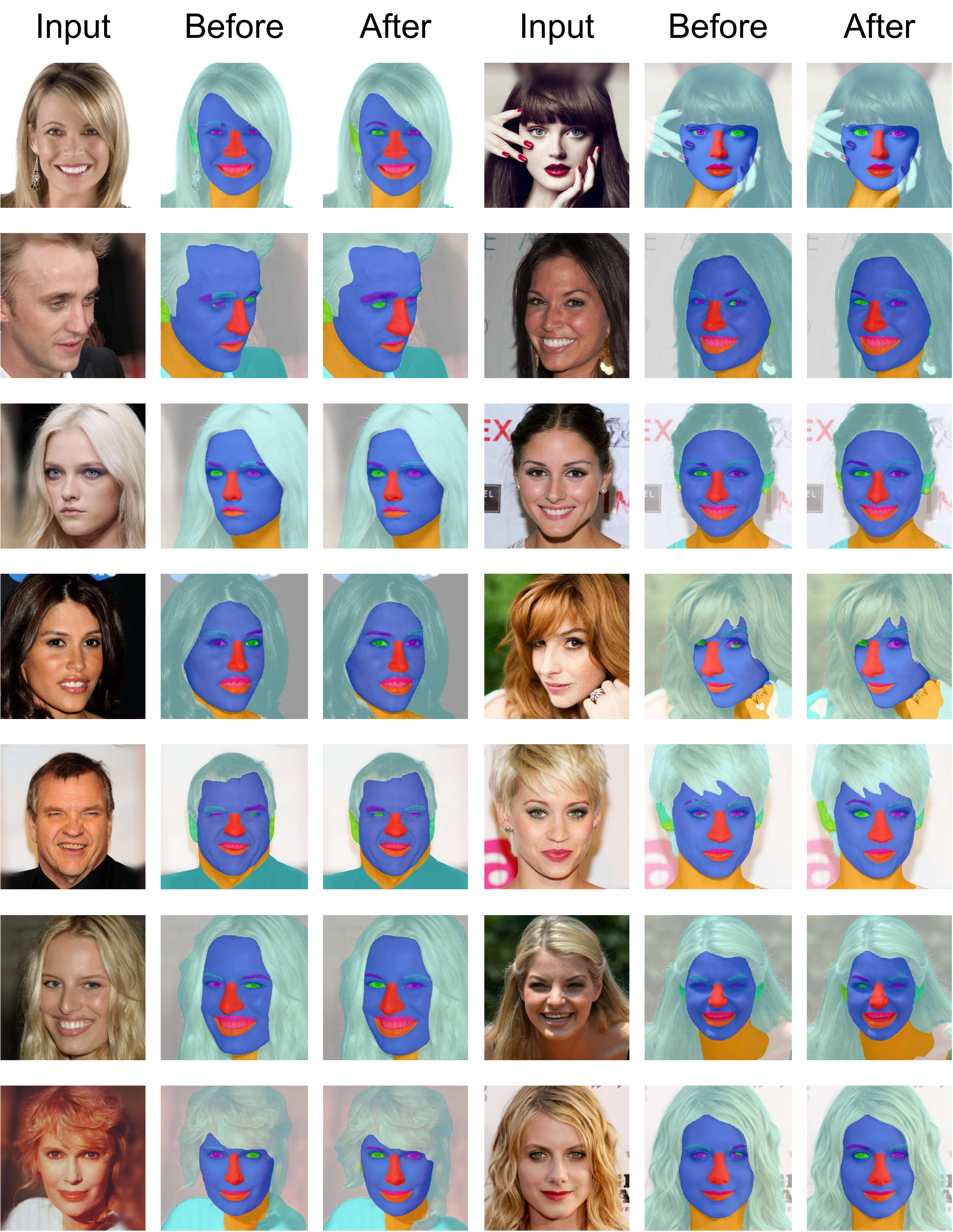}
\end{figure*}

\begin{figure*}
    \centering
    \includegraphics[trim=0cm 0cm 0cm 0cm, clip, width=\linewidth]{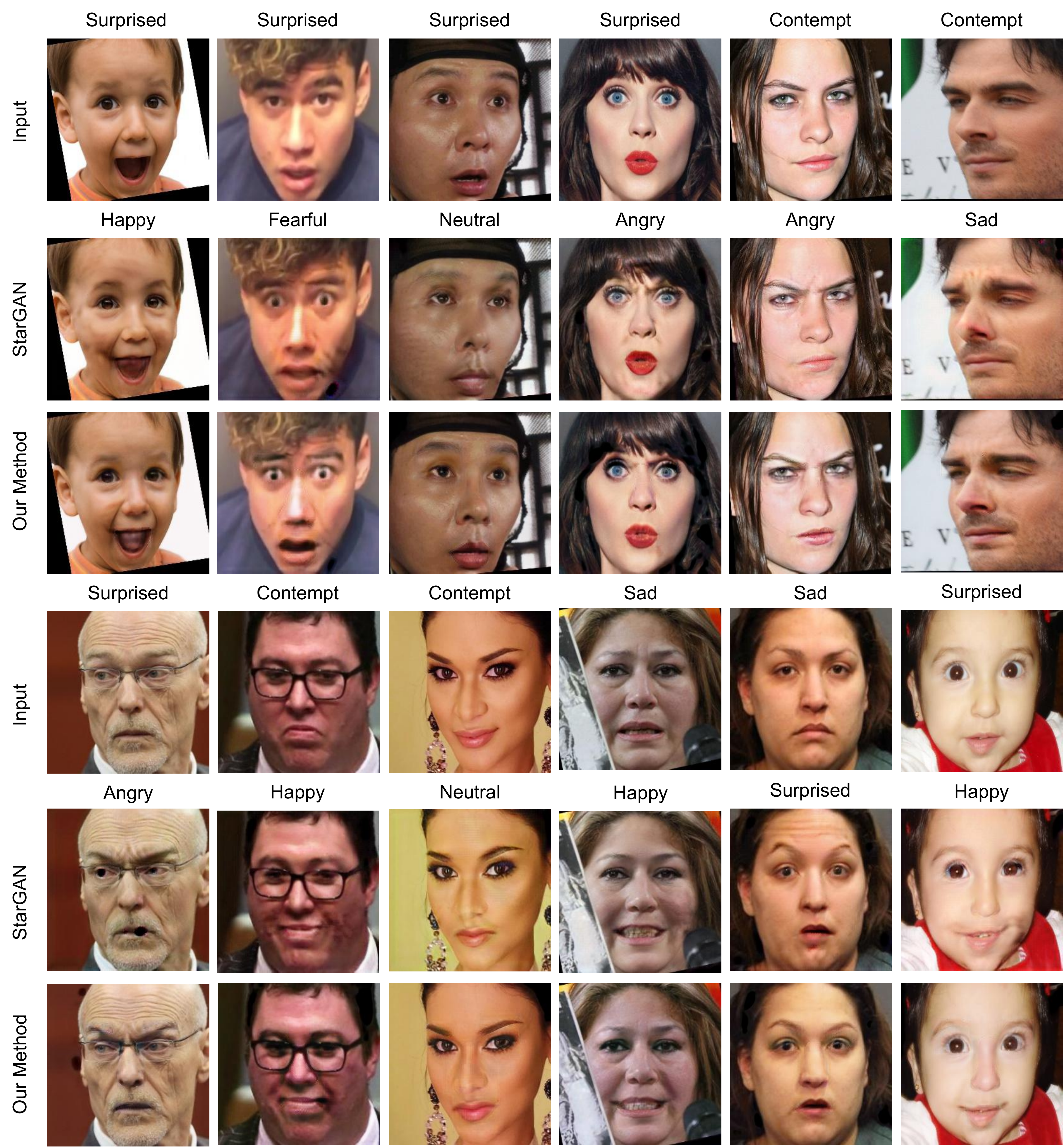}
\end{figure*}

